%% file: AnonymousSubmission2027.tex
\documentclass[letterpaper]{article} 
\usepackage{aaai2027}  
\usepackage[hyphens]{url}  
\usepackage{graphicx} 
\usepackage{natbib}  
\usepackage{caption} 
\usepackage{algorithm}
\usepackage{algorithmic}

\usepackage{newfloat}
\usepackage{listings}
\DeclareCaptionStyle{ruled}{labelfont=normalfont,labelsep=colon,strut=off} 
\floatstyle{ruled}
\newfloat{listing}{tb}{lst}{}
\floatname{listing}{Listing}

\usepackage{booktabs}

\usepackage{cuted}
\usepackage{amsmath}
\usepackage{amssymb}
\newcommand{\name}{ES3D}

\title{ES3D: Component-Aware 3D Editing via 3D Semantic Embedding}
\author{
    Xuancheng Jin\textsuperscript{\rm 1}, Rengan Xie\textsuperscript{\rm 1}, Jiayuan Lu\textsuperscript{\rm 1}, Wenting Zheng\textsuperscript{\rm 1},\\
    Rui Wang\textsuperscript{\rm 1}, Lincheng Li\textsuperscript{\rm 2}, Yingfeng Chen\textsuperscript{\rm 2}, Yuchi Huo\textsuperscript{\rm 1}\corresponding
}
\affiliations{
    \textsuperscript{\rm 1}State Key Laboratory of CAD\&CG, Zhejiang University\\
    \textsuperscript{\rm 2}NetEase Fuxi AI Lab
}

\begin{document}

\maketitle

\input{sec/0_abstract}

\input{sec/1_introduction}
\input{sec/2_related_work}
\input{sec/3_method}
\input{sec/4_experiment}
\input{sec/5_misc}

\bibliography{aaai2027}


\end{document}

%% file: sec/0_abstract.tex
\begin{abstract}

Existing 3D editing methods have made notable progress in controllability, yet they remain limited in several important ways. 
Most approaches rely on text-driven editing, which struggles to express fine-grained visual changes intended by the user. 
Moreover, many methods require manually supplied 3D masks or introduce unintended changes to regions that should remain untouched. 
These limitations largely arise from the absence of fine-grained semantic understanding, making it difficult for existing models to retrieve or modify specific 3D components.

We introduce \name, a framework that embeds semantics directly into 3D space, enabling component-aware retrieval and editing of a 3D asset conditioned on multiple local reference images and optional text queries.
We first construct a 3D semantic embedding by projecting multi-view semantic features into the voxelized space of the asset.
We then perform 3D component retrieval by computing feature similarity between the 3D semantic embedding and the semantic embeddings of image or text queries.
For editing, we employ a pretrained 3D generative model with an inpainting mechanism to modify the retrieved components guided by user-provided images while preserving the rest of the asset.
Overall, ES3D is a 3D editing framework that retrieves editable regions based on semantic cues and uses multiple images as conditions.
Extensive experiments demonstrate that \name{} produces geometrically consistent and semantically coherent edits, enabling robust image-based and text-assisted control for 3D editing.
\end{abstract}

%% file: sec/1_introduction.tex


\section{Introduction}
\label{sec:intro}

Recent advances in 3D generation have achieved remarkable progress in producing high-fidelity geometry and appearance from text or image inputs.
However, editing an existing 3D asset remains a much more challenging task.
Accurate and controllable 3D editing is crucial for real-world applications such as content creation, animation, and virtual asset design, where users often wish to modify specific parts of a model without affecting the rest.

Existing 3D editing approaches fall broadly into two categories, each with notable limitations.
The first type requires explicit masks to indicate the regions to be modified \cite{meshEditing, A3D, progressive3D, meshEditing}.
While these methods can preserve the untouched areas, they rely heavily on manual intervention.
The second type performs 2D-based modifications using diffusion models and then propagates the changes back to 3D space \cite{pro3d, preditor3d, focalDream, nano3d}.
Although this requires no manual 3D mask, it often introduces global inconsistencies and alters regions that should remain unchanged.
Both categories are predominantly text-conditioned, which further restricts their ability to express fine-grained visual intent or support image-driven edits.

\begin{figure}[H]
  \centering
  \includegraphics[width=\columnwidth]{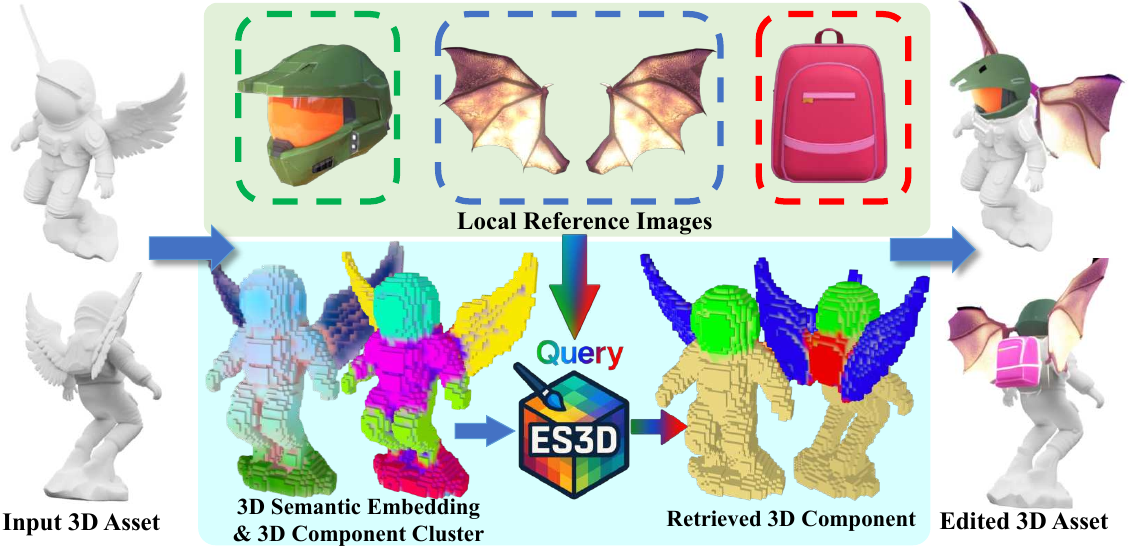}
  \caption{\name{} is capable of editing 3D asset conditioned on multiple local reference images.}
  \label{fig:teaser}
\end{figure}

This analysis reveals that \emph{automatically identifying which 3D regions should be modified is the key to edit 3D assets} without relying on explicit 3D masks or uncontrolled propagation.
Consequently, we propose ES3D, a framework that embeds semantics directly into 3D space, enabling automatic component retrieval and precise 3D editing guided by semantic cues from images or optional text.
Given a 3D mesh and one or more local reference images with optional text, ES3D identifies and modifies the corresponding 3D components in a semantically consistent manner.
To achieve this, we leverage DINOv3\cite{dinov3}, an image encoder with strong semantic understanding, to align multi-view 2D semantics with the 3D geometry of a given asset.
Specifically, given a 3D mesh, we voxelize it and render multiple views as input images, which are then encoded by DINOv3 to extract semantic features.
The extracted features are then projected onto the voxels to obtain a 3D semantic embedding.
However, since the DINOv3 feature maps have limited spatial resolution, directly projecting them into voxels can blur semantic boundaries.
To mitigate this, we propose Semantic Feature Enhancement Module, producing sharper and more detailed 3D semantic embeddings.

Based on this semantic embedding, we can retrieve 3D regions using either image or text queries via feature similarity, as we use the text-aligned version of DINOv3, which maps visual and textual inputs into a shared semantic space.
However, voxel-level retrieval tends to produce noisy and unstable boundaries, especially near component edges. 
Therefore, we first cluster the voxel embeddings into semantically coherent components and then perform similarity search at the component-level rather than the voxel-level. 
This cluster-level retrieval suppresses boundary noise and yields clean, consistent 3D masks.


After retrieving the 3D components, we adopt the pretrained 3D generative model TRELLIS~\cite{trellis} as our editing backbone. 
Following the insight of Fuse3D~\cite{fuse3d}, we incorporate multi-masked-image feature extraction strategy and use the input local images as additional conditions to guide the edit.

A key challenge here is that these inputs are local reference images, whereas TRELLIS is trained to colorize voxels under whole-view condition image. 
As a result, directly encoding a local crop fails to convey its correct spatial placement within the global 3D structure.
To address this, we introduce an image stacking strategy to provide explicit spatial cues to the model.
Before encoding each local reference image, we paste it onto a selected rendered view at the corresponding location, then encode the stacked image and extract the tokens associated with the pasted region. 
These tokens are concatenated with tokens extracted from the remaining rendered views to form the unified conditional tokens used for editing.
Based on these tokens, editing is performed in two stages.
We first inpaint the voxel grid using the unified condition tokens together with a refined inpainting mask to update only the selected component while keeping other areas fixed, and then inpaint the appearance features using the same tokens to obtain the edited 3D asset.

\begin{figure}
  \centering
  \includegraphics[width=.95\columnwidth]{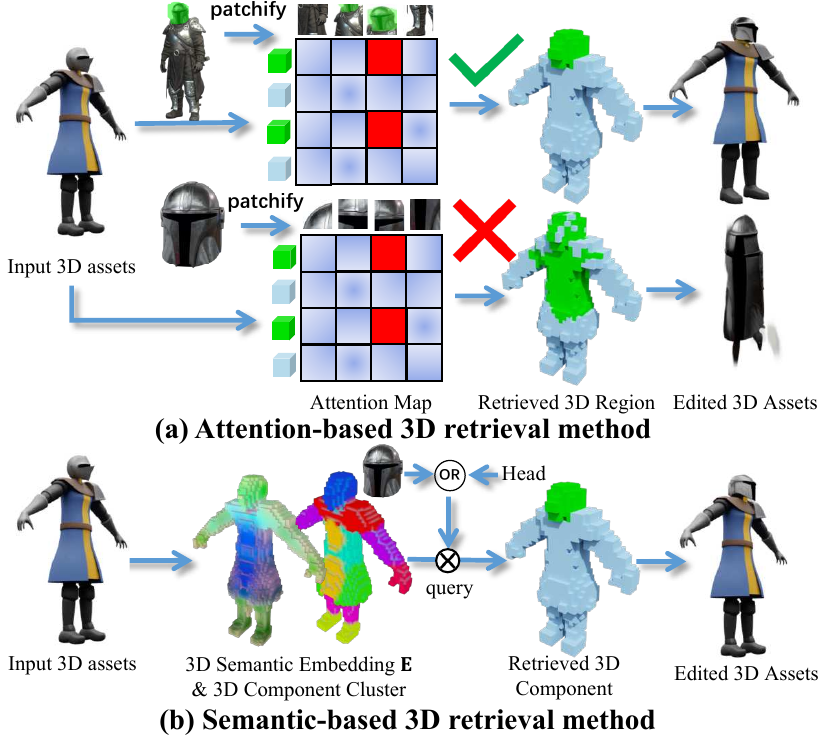}
  \caption{
    Difference between Fuse3D and ES3D in 3D retrieval.
    \textbf{(a)} Fuse3D retrieves 3D regions by leveraging cross-attention module inside pretrained image-to-3D models, 
    but this mechanism breaks when the reference image only covers a partial view of the full object.
    \textbf{(b)} ES3D constructs a voxel-level semantic embedding and performs cluster-level 3D component retrieval, enabling robust localization from either local reference images or text queries.
  }
  \label{fig:fuse3d_diff}
\end{figure}

Compared to related work such as Fuse3D, which performs 3D retrieval based on attention mechanism, our method retrieves 3D components based on explicit semantic cues.  
This overcomes the main limitation illustrated in Fig.~\ref{fig:fuse3d_diff}, where Fuse3D fails when the reference image covers only a partial region of the object.
Overall, the main contributions of this work can be summarized as follows.
\begin{itemize}
\item To the best of our knowledge, ES3D is the first framework to provide explicit 3D semantic embedding information for 3D asset editing, enabling accurate retrieval of editable components and facilitating component-aware 3D editing.
\item We present a novel method that enables precise 3D asset editing driven by local reference images, offering finer visual control than text-based approaches.
\item We propose an image stacking strategy that provides spatial context for local reference images, bridging the gap between local conditioning at inference and the global setup used during training.
\end{itemize}

%% file: sec/2_related_work.tex
\section{Related Work}

\subsection{2D Editing}


Early diffusion-based \cite{ddpm} editing approaches typically relied on user-provided 2D masks to define editable regions, followed by inpainting to generate coherent content consistent with the surrounding context~\cite{repaint,blendedDiffusion,glide}.
Although effective for localized editing, these methods depend on manually specified masks, which are not always available.

Subsequent approaches introduced inversion-based strategies that reconstruct the latent representation of an image and perform editing through diffusion guidance~\cite{sdedit, nullText, ledit++, turboEdit}. 
While effective, these methods often lack spatial precision and tend to alter unintended areas. 

Recent works further explore attention-level control to preserve spatial layout and semantic consistency during editing. 
Those methods \cite{p2p, plugPlay, masactrl} achieve controllable edits by manipulating cross- and self-attention maps, ensuring that object positions and scene composition remain consistent while altering visual attributes or concepts.

\subsection{3D Object Generation}

The field of 3D generative modeling has seen remarkable progress in recent years, driven by diffusion-based frameworks that enable high-fidelity synthesis across various 3D representations, including point clouds~\cite{pointE}, implicit fields~\cite{SDFusion, shapeE, diffusionSDF}, and explicit meshes~\cite{GET3D}.
These methods directly model the distribution of 3D shapes via diffusion, enabling high-quality outputs.

Building on diffusion priors, methods such as DreamFusion~\cite{dreamFusion} employ Score Distillation Sampling (SDS) to lift 2D generation capabilities into 3D.
Although effective, these optimization-based approaches are computationally expensive and often suffer from instability, including view-inconsistent “Janus” artifacts.
To address these issues, recent works like LRM~\cite{lrm,lgm,ldm} propose end-to-end transformer-based models trained directly on large-scale 3D data, removing the dependence on noisy gradients from 2D priors and achieving more stable and efficient generation.

Most recently, the TRELLIS framework~\cite{trellis} represents a 3D-native generative paradigm that adopts a two-stage pipeline, where a VAE~\cite{vae} maps 3D assets into a latent space and diffusion-based generation is performed within this space.
This structured design leads to stable, high-fidelity results and has inspired several extensions~\cite{ultra3d,direct3d2s}.

\subsection{3D Object Editing}

Compared to 2D image editing, maintaining spatial consistency in 3D editing remains highly challenging.
Recent diffusion-based approaches for 3D editing can be broadly categorized into two paradigms.
The first relies on explicit spatial masks to specify editable regions~\cite{meshEditing, A3D, progressive3D, Instant3dit}, offering precise control but requiring manual mask.
The second performs editing in 2D and propagates the changes to 3D space~\cite{focalDream, instructNerf, pro3d, preditor3d}, which avoids manual 3D masks but leads to unstable reconstructions due to disrupted geometric coherence.
Some methods further combine these paradigms by introducing spatial constraints into 2D-guided editing~\cite{MagicClay}.
Notably, most existing approaches are primarily text-driven, limiting fine-grained visual control. Although a few methods incorporate image-based conditions~\cite{pixbrush}, their effects are restricted to texture synthesis without modifying geometry.

Among prior works, Nano3D~\cite{nano3d} is the closest to ours, as it also adopts TRELLIS as the generative backbone for 3D editing.
However, Nano3D is purely text-driven and infers editable regions by propagating 2D edits into 3D without explicit semantic guidance, which limits fine-grained visual control and often leads to unstable localization.
In contrast, our method embeds semantic features directly into 3D space and retrieves editable components via semantic similarity, enabling precise region localization and multi-image–guided editing.

%% file: sec/3_method.tex
\section{Method}

\begin{figure*}[!htp]
    \centering
    \includegraphics[width=\textwidth]{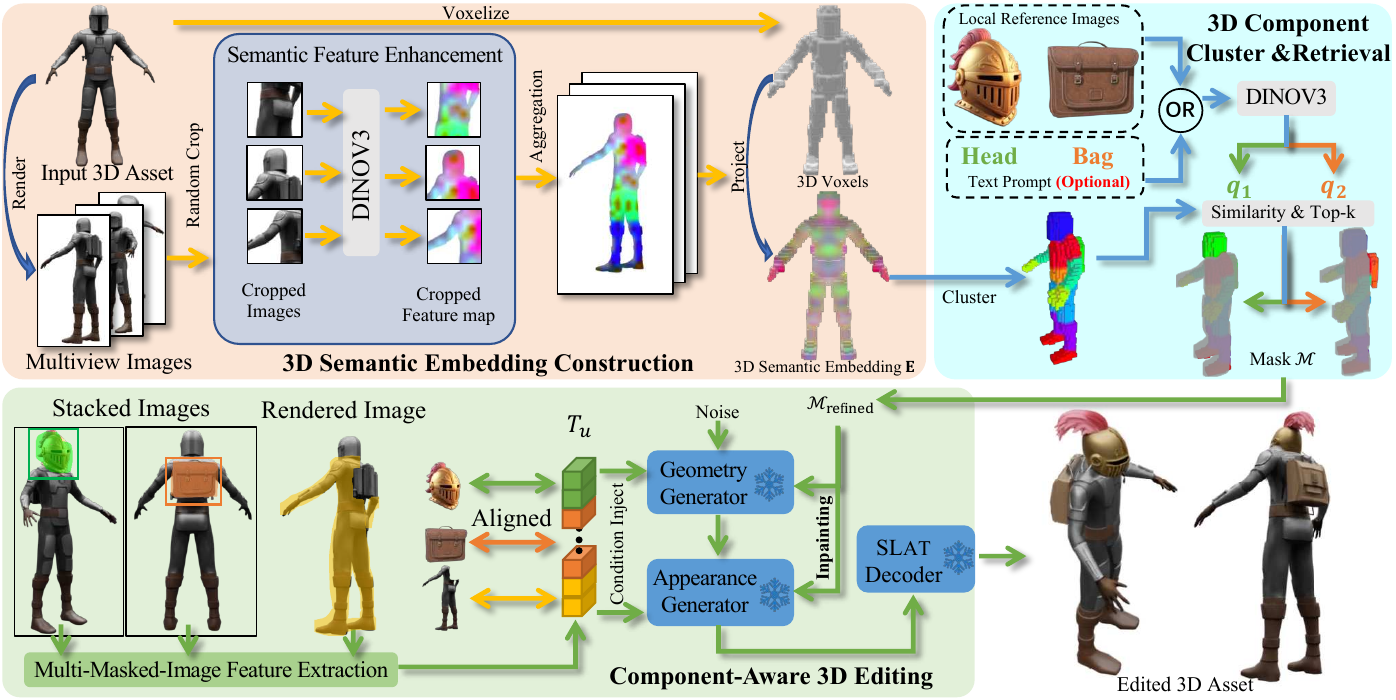}
    \caption{{
Given a 3D asset along with multiple local reference images and optional text, we first construct a 3D semantic embedding on the voxel grid.
We then perform clustering on this embedding and retrieve the target component using image or text queries.
Finally, a unified conditional token, produced by multi-masked-image feature extraction module, is injected into a pretrained 3D generative model together with a refined inpainting mask to obtain the final edited 3D asset.
    } 
    }
   \label{fig:overview}
\end{figure*}

We introduce \textbf{ES3D}, a framework that enables component-aware 3D editing conditioned by image prompts. 
Fig~\ref{fig:overview} shows an overview of our method.
We begin by introducing the preliminary background in Sec.~\ref{sec:preliminary}. 
Sec.~\ref{sec:3dConstruct} then presents our 3D Semantic Embedding Construction. 
Next, we describe how we retrieve 3D components given semantic cues in Sec.~\ref{sec:3dRetrieve}. 
Finally, we illustrate how the pretrained generative model is used to perform component-aware 3D editing in Sec.~\ref{sec:3dEdit}.

\subsection{Preliminary}
\label{sec:preliminary}

\paragraph{TRELLIS}
We adopt TRELLIS~\cite{trellis} as the backbone of our editing framework.
For a 3D asset $\mathcal O$, TRELLIS is capable of encoding a 3D assets into a \textsc{SLat} representation, which is defined as:

{\small
\begin{equation}
    \boldsymbol{Z} = \{(\boldsymbol{z}_i,\boldsymbol{p}_i)\}_{i=1}^{L},\quad 
    \boldsymbol{z}_i\in\mathbb{R}^C, \ \boldsymbol{p}_i\in \{0, 1,\ldots, N-1\} ^3, 
    \label{eq:slate}
\end{equation}
}
where $\boldsymbol{p}_i$ denotes the index of an active voxel in 3D grid, $\boldsymbol{z}_i$ is a local latent attached to voxel $\boldsymbol{p}_i$, $L$ is the total number of active voxels, $C$ is the channel dimension and $N$ is the spatial length of 3D grid.
Based on \textsc{SLat} representation, TRELLIS generates the 3D asset in a two-stage pipeline.
In the first stage, it generates a low-resolution feature field $\boldsymbol{S}\in\mathbb{R}^{D\times D\times D \times C_\mathrm{S}}$, which is then decoded into $\{\boldsymbol{p}_i\}_{i=1}^{L}$.
In the second stage, TRELLIS generates the corresponding latent features $\{\boldsymbol{z}_i\}_{i=1}^{L}$ for these voxels.
The complete \textsc{SLat} representation $\boldsymbol{Z}$ is then processed through specialized decoders to produce the final 3D asset $\mathcal{O}$.

\paragraph{Fuse3D}

We are inspired by the multi-masked-image feature extraction strategy of Fuse3D~\cite{fuse3d} and incorporate it into our pipeline to perform inpainting within the TRELLIS generation process for 3D editing. 
Fuse3D introduces a mechanism for combining visual tokens from multiple images as condition for TRELLIS, 
allowing the model to fuse visual cues from different images while supporting region-level control for each image. 
Concretely, for each image paired with a mask, DINOv2 is used to encode the image, and the tokens corresponding to the masked region are extracted. 
These extracted tokens from multiple images are concatenated to form unified condition tokens, 
which serve as the generation condition for TRELLIS and enable the fusion of multiple regions across images.

\subsection{3D Semantic Embedding Construction}
\label{sec:3dConstruct}


The goal of this stage is to construct a 3D semantic embedding that allows image or text queries to perform component-level retrieval within 3D space. 
To achieve this, we project multi-view semantic features extracted by a pretrained DINOv3 \cite{dinov3} encoder $f_{\text{DINO}}$ into the 3D space of the given asset, as DINOv3 features are well aligned with high-level semantics.

Specifically, given an input mesh $\mathcal{M}$, we first render $N$ images $\{I_i\}_{i=1}^{N}$ under known cameras $\{\mathcal{C}_i\}_{i=1}^{N}$ and obtain their feature maps $F_i$ using $f_{\text{DINO}}$. 
However, due to the limited resolution of DINOv3 feature maps, directly projecting them into 3D results in blurred semantics, especially around small or thin structures.
To alleviate this issue, we enhance the encoding process by introducing Semantic Feature Enhancement, inspired by the multi-crop training strategy \cite{multi_crop_train} of DINO \cite{dino}, where we adopt a similar multi-crop encoding scheme at inference time to obtain denser semantic features.
For each rendered image $I_i$, multiple random crops $\{I_i^k\}_{k=1}^{K}$ are encoded by $f_{\text{DINO}}$ and aggregated back to their original coordinates, yielding a higher-resolution feature map $\tilde{F}_i$:

\begin{equation}
\tilde{F}_i = \text{Aggregate}\!\left(\{f_{\text{DINO}}(I_i^k)\}_{k=1}^{K}\right) \in \mathbb{R}^{H_f \times W_f \times D}.
\end{equation}

Here, $\text{Aggregate}(\cdot)$ denotes a feature-space fusion operator that aligns crop-level DINOv3 feature maps onto a higher-resolution feature grid of size $(H_f, W_f)$ using bilinear interpolation, followed by average pooling over overlapping feature regions.

After obtaining high-resolution feature maps for each rendered view, we voxelize the input mesh to ensure consistency with the downstream TRELLIS pipeline, which operates on voxel-based representations.
The resulting voxels are denoted as $\{\boldsymbol{p}_i\}_{i=1}^{L}$, where $L$ is the number of voxels. 
Each voxel receives its semantic feature by back-projecting the aggregated multi-view features $\{\tilde{F}_i\}_{i=1}^{N}$ according to the known camera parameters $\{\mathcal{C}_i\}_{i=1}^{N}$. 
In this way, we obtain a 3D semantic embedding volume:

\begin{equation}
\mathbf{E} = \{\boldsymbol{e}_i\}_{i=1}^{L} \in \mathbb{R}^{L \times D},    
\end{equation}

where $D$ is the dimension of the semantic feature. 
This embedding provides component-level semantic representations that serve as the foundation for subsequent editing.

\subsection{3D Component Cluster \& Retrieval}
\label{sec:3dRetrieve}


After constructing the 3D semantic embedding, our retrieval stage identifies the component to be edited, conditioned on $M$ local reference images.
In cases where the intended editing region does not semantically align with the image, the user can additionally provide a text query to locate the corresponding 3D component, while the image remains the appearance reference for subsequent editing.

Specifically, for each input query $Q_i$, which can be either an image or text, 
we encode it into the same semantic space as the 3D embedding using the DINOv3 encoder:
\begin{equation}
\mathbf{q}_i = f_{\text{DINO}}(Q_i) \in \mathbb{R}^D.
\end{equation}
This allows direct similarity comparison between $\mathbf{q}_i$ and voxel-level features $\boldsymbol{e}_i$. 
A naive approach computes voxel-wise cosine similarity and thresholds it to form a 3D mask. 
However, such voxel-level retrieval often produces noisy boundaries due to local semantic ambiguity.

To alleviate this issue, we perform clustering over the 3D embedding space $\mathbf{E}$ to obtain semantically consistent components 
$\{P_k\}_{k=1}^{K_c}$ (detailed in Appendix), where each component $P_k$ groups voxels sharing similar features and $K_c$ is a user-defined parameter. 
For each component, we compute the mean feature:

\begin{equation}
\bar{\mathbf{E}}_k = \frac{1}{|P_k|}\sum_{\mathbf{p}_i \in P_k} \boldsymbol{e}_i.
\end{equation}
We then evaluate the cosine similarity between the query embedding and each component mean feature:

\begin{equation}
s_k^i = \frac{\bar{\mathbf{E}}_k \cdot \mathbf{q}_i}{\|\bar{\mathbf{E}}_k\| \, \|\mathbf{q}_i\|},
\end{equation}
and select the top-$\mathcal{K}$ components with the highest similarity as the retrieved regions, with $\mathcal{K}$ defined by the user.
The resulting mask $\mathcal{M}_i\in\{0,1\}^{L}$ indicates the component-level 3D voxels that correspond to the $i$-th input query.

\subsection{Component-Aware 3D Editing}
\label{sec:3dEdit}

\paragraph{Condition Tokens Construction}


After retrieving the target 3D component, we employ the pretrained TRELLIS generative model to inpaint this region while keeping the remaining areas unchanged. 
A key challenge in this process lies in constructing appropriate condition tokens. 
We observe that directly encoding the local reference image as inpainting conditions leads to severe degradation in generation quality (see Fig~\ref{fig:ablation_2}). 
Inspired by Fuse3D~\cite{fuse3d}, we attribute this limitation to the lack of spatial context during image encoding.
Without positional cues or cross-image attention, the model cannot infer where the given local image fits within the global 3D structure (e.g., encoding only a head provides no information that it belongs to a body).

To address this issue, we introduce a simple yet effective \emph{image stacking} strategy. 
Specifically, before encoding each local reference image, we paste it onto a rendered view of the object at the corresponding position, providing spatial cues to the model. 
Notably, this stacking does not need to be pixel-accurate; an approximate spatial alignment is sufficient to convey the spatial information for encoding.

Following the multi-masked-image feature extraction module in Fuse3D, we encode the stacked images and extract the tokens corresponding to the stacked regions, which serve as the condition for the edited component. 
To ensure stable generation, we also construct condition tokens for the regions that should remain unchanged. 
Since the 3D regions to be preserved are already known, we render an additional reference view where those areas are masked, encode it, and extract the corresponding tokens. 
Tokens from edited and preserved regions are concatenated to form the unified condition tokens $T_u$ for inpainting (detailed in Appendix).

\paragraph{3D Assets Editing}


Given the unified condition tokens $T_u$, this stage aims to edit the retrieved 3D components produced in Sec~\ref{sec:3dRetrieve} by performing inpainting. 
We apply inpainting to both stages of the TRELLIS model, including geometry and appearance generation.

Before inpainting, we construct a unified mask $\mathcal{M}$ by taking the union of all retrieved masks $\mathcal{M} = \bigcup_{i=1}^{M}\mathcal{M}_i,$
where each $\mathcal{M}_i$ corresponds to the region retrieved for the $i$-th query.  
In the geometry generation stage, TRELLIS operates on a low-resolution coordinate field 
$\boldsymbol{S}\in\mathbb{R}^{D\times D\times D \times C_\mathrm{S}}$. 
To align $\mathcal{M}$ with the resolution of $\boldsymbol{S}$, we downsample it to obtain 
$\mathcal{M}_d$ and derive a complementary keep mask $\mathcal{K}_d$.  
Unlike traditional 2D inpainting, which only considers visible pixels, 3D geometry generation must also account for unoccupied spatial regions in the dense voxel field.  
Directly regenerating $\mathcal{M}_d$ or preserving $\mathcal{K}_d$ can hinder new geometry growth in empty regions or introduce structural distortions.
To address this, we apply a 3D binary dilation to $\mathcal{M}_d$ to obtain a refined inpainting mask $\mathcal{M}_{\text{refined}}$ (detailed in Appendix), which enables proper geometric growth within the edited regions while avoiding unintended modifications to the preserved areas.

During generation, the unified condition tokens are incorporated into the diffusion sampling through an inpainting strategy.
Specifically, at each time step $t$, we manually perturb the known latent $\boldsymbol{S}^{\text{known}}$ using a forward process \cite{flow} to get the unknown latent:
\vspace{-0.5em}
\begin{equation}
\boldsymbol{S}_{t}^{\text{known}} = (1-t)\boldsymbol{S}^{\text{known}} + t\epsilon,
\end{equation}
\vspace{-0.5em}
where the $\epsilon$ are random noises. The update rule is:

\vspace{-0.5em}
\begin{equation}
\boldsymbol{S}_{t+1} = \mathcal{M}_{\text{refined}} \odot \boldsymbol{S}_{t}
+ (1 - \mathcal{M}_{\text{refined}}) \odot \boldsymbol{S}_{t}^{\text{known}},
\end{equation}
where $\odot$ denotes element-wise multiplication, $\boldsymbol{S}_{t}$ represents the predicted latent at time step $t$.
After geometry decoding, minor artifacts may appear in preserved regions, which we mitigate via a simple post-processing step based on spatial edit localization (see Appendix).

In the second stage, which generates appearance features without a VAE bottleneck, we first calculate the inpainting mask $\mathcal{M}_{r}^A$ by identifying voxels that remain unchanged in the geometry inpainting stage.
Using inpainting mask $\mathcal{M}_{r}^A$ and the same conditional tokens, we perform appearance inpainting to obtain the final edited 3D asset.

%% file: sec/4_experiment.tex
\begin{figure*}[!htp]
  \centering
  \includegraphics[width=\textwidth,height=.35\textheight,keepaspectratio]{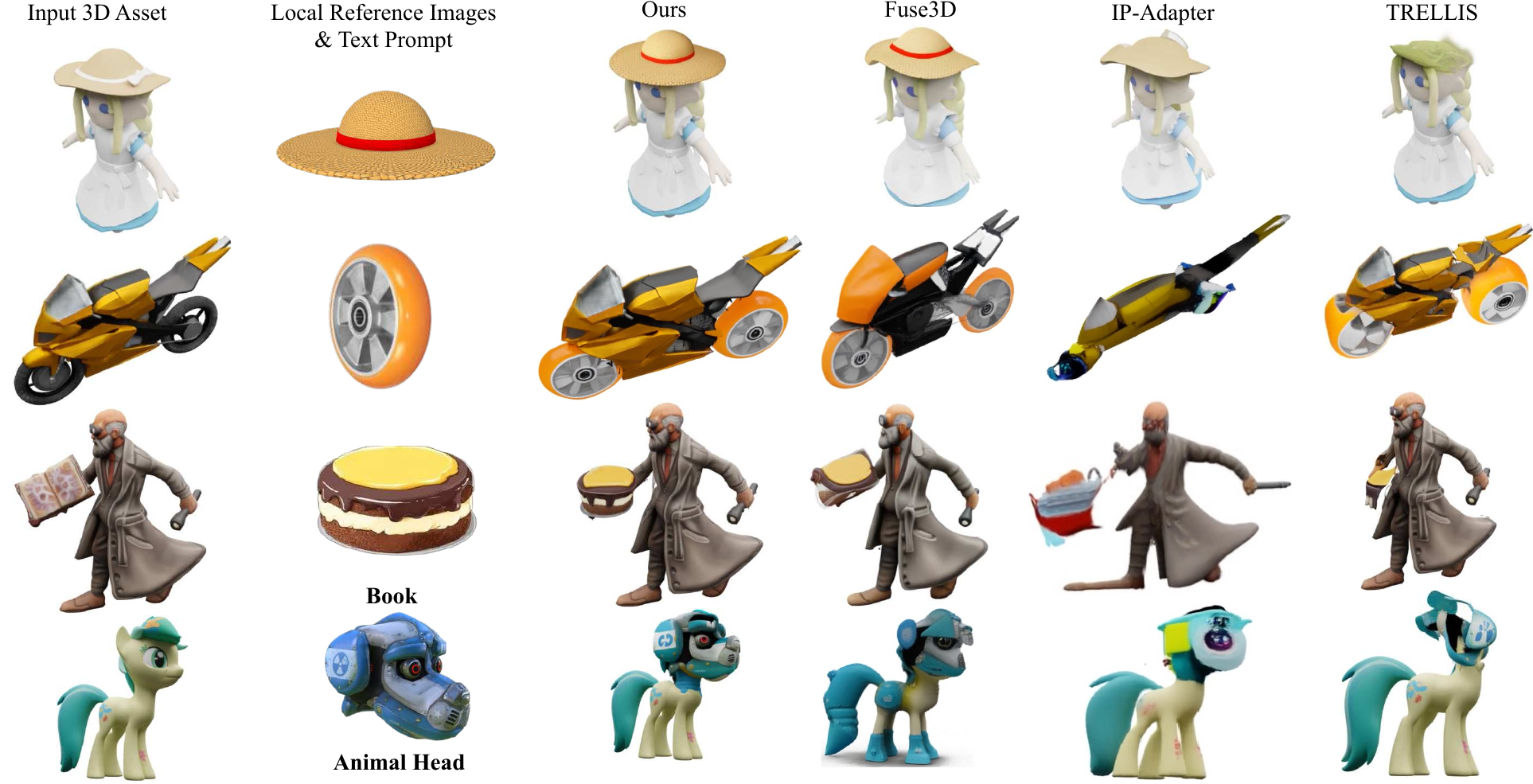}

  \caption{
  Editing quality comparison with previous methods.
  }
  \label{fig:compare_1}
\end{figure*}

\section{Experiments}

\subsection{Baselines}

Since there are \textbf{\emph{no existing 3D editing methods that are directly controlled by reference images}}, selecting fair and comparable baselines is highly challenging. 
To enable meaningful comparison, we evaluate our method against three approaches:  
\textbf{(1) Fuse3D}~\cite{fuse3d}, which fuses multiple reference images to jointly control 3D generation;  
\textbf{(2) IP-Adapter}~\cite{ipAdapter}, a 2D image editing model that conditions on reference images. 
For fair comparison, we extend its results to 3D by applying a 2D-to-3D reconstruction pipeline, TRELLIS; and  
\textbf{(3) TRELLIS}~\cite{trellis}, a 3D generative model that uses manually defined 3D masks for inpainting.  
Implementation details of these baselines are provided in the Appendix. 
For completeness, the Appendix also includes comparisons with state-of-the-art text-controlled 3D editing and 3D localization methods, as well as a limitation analysis.

\begin{figure}
  \centering
  \includegraphics[width=\columnwidth]{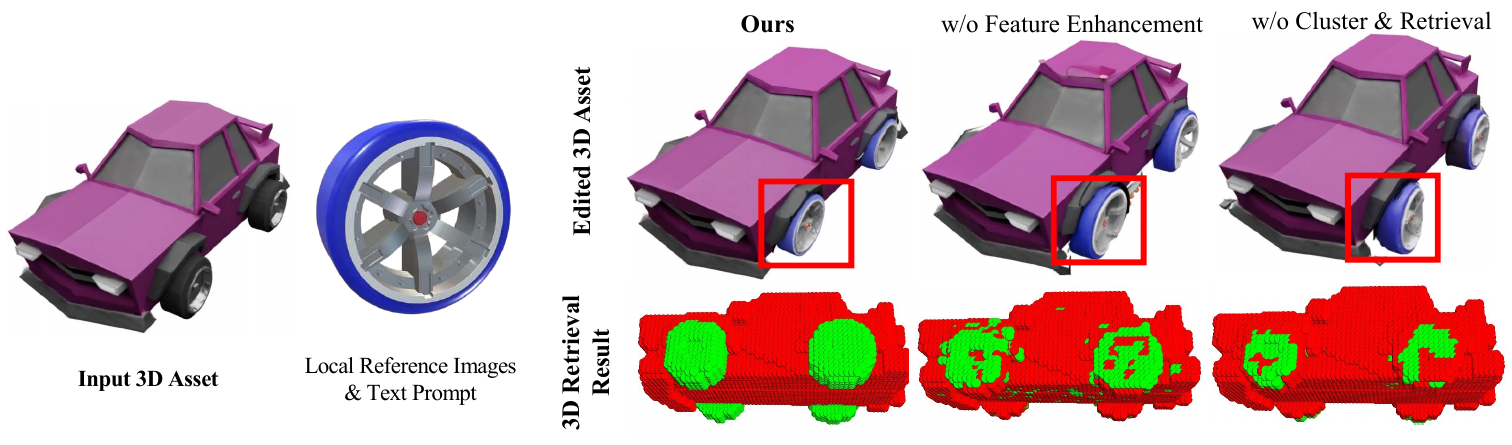}
  \caption{
  Ablation study of Semantic Feature Enhancement Module and 3D Component Cluster \& Retrieval Module.
  }
  \label{fig:ablation_1}
\end{figure}

\vspace{-0.5em}

\subsection{Comparison Results}

\paragraph{Qualitative Evaluation}

As shown in Fig.~\ref{fig:compare_1}, our method produces the most precise and coherent edits among all compared methods. Fuse3D can integrate multi-view features, but cannot preserve unedited regions or substantially modify geometry; for example, the shape of a hat remains unchanged while only its appearance is altered. Since Fuse3D also struggles when the reference image covers only part of the voxel grid (Fig.~\ref{fig:fuse3d_diff}), we provide it with manually added global context, as detailed in the Appendix. IP-Adapter relies on 2D editing followed by 2D-to-3D reconstruction, which introduces inconsistencies in preserved regions. TRELLIS with inpainting can better preserve unedited areas, but requires manually specified 3D masks and produces weaker edits due to limited spatial information, i.e., no image stacking. In contrast, ES3D enables component-aware retrieval and editing, accurately localizing target regions, producing high-quality edits, and preserving unrelated areas.

\paragraph{Quantitative Evaluation}

\input{tables/compare_quantity}
\input{tables/gpt_eval}
\input{tables/userStudy}


We evaluate our method and all baselines using CLIP, region-specific CLIP, and ImageReward for edit quality, CD$_{\text{keep}}$ for preservation of unedited regions, and 3D BBox IoU for target-region localization accuracy. CD$_{\text{keep}}$ measures the Chamfer distance between the unedited regions before and after editing. 
As shown in Table~\ref{tab:quant_metrics}, ES3D achieves the best performance on all metrics.
In addition, following Fuse3D, we assess our method with GPT-Eval3D \cite{gptEval} across visual quality, editing accuracy, and preservation consistency, and further validate it with a user study (details in Appendix).
As shown in Table~\ref{tab:GPTcompare} and Table~\ref{tab:userStudy}, ES3D consistently outperforms all baseline.

\begin{figure*}[t]
  \centering
  \includegraphics[width=\textwidth,height=.42\textheight,keepaspectratio]{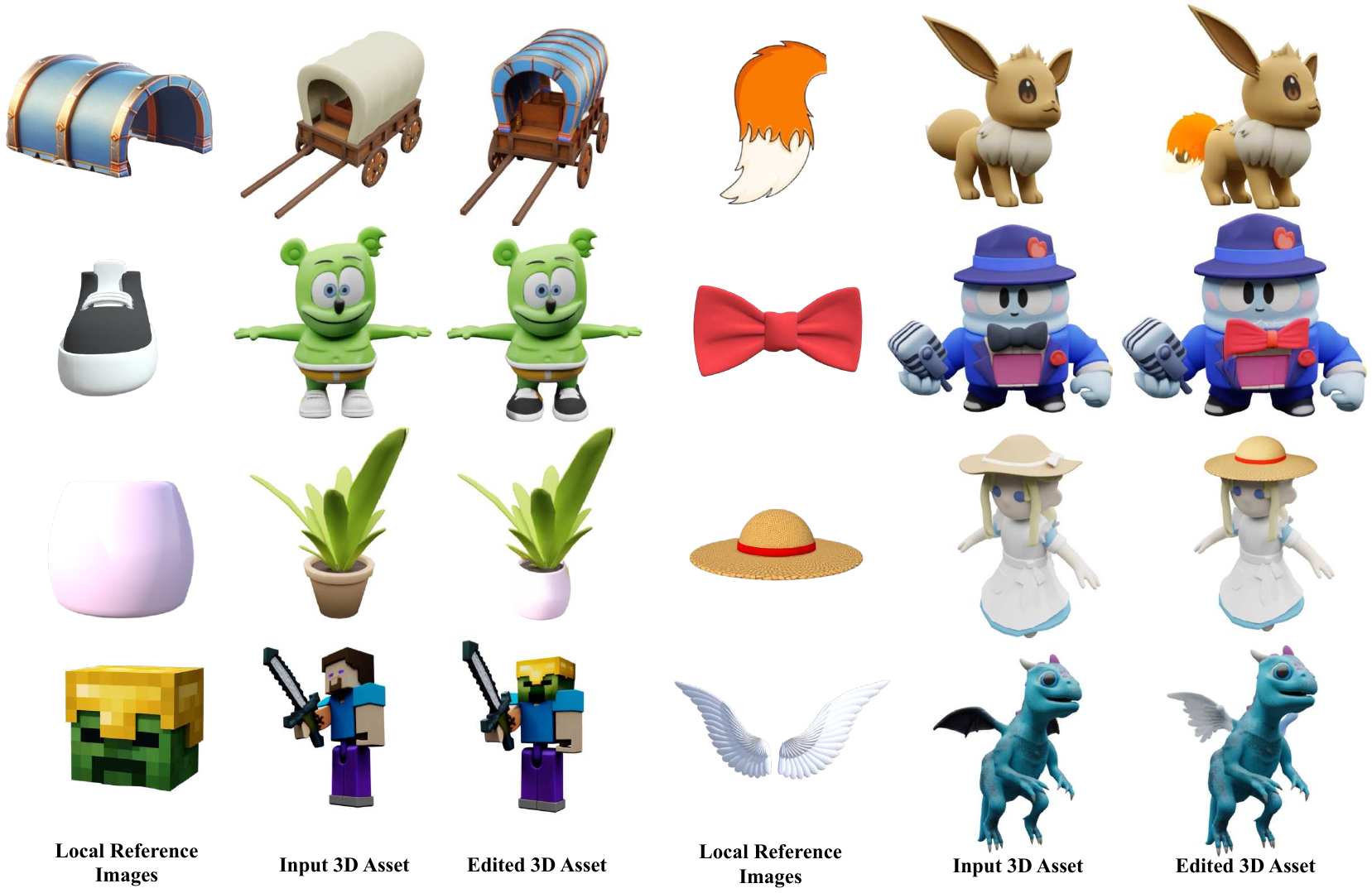}
  \vspace{-1em}
  \caption{
  More results of edited 3d assets.
  }
  \label{fig:mainCompare}
\end{figure*}

\begin{figure}
  \centering
  \includegraphics[width=\columnwidth]{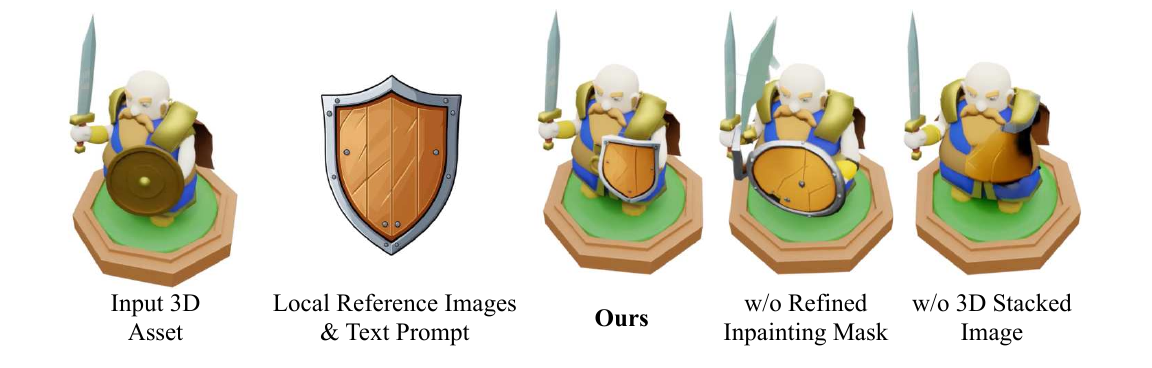}
  \caption{
  Ablation study of Refined Inpainting Mask and Stacked Image strategy.
  }
  \label{fig:ablation_2}
\end{figure}

\subsection{Ablation Study}

\input{tables/ablation_quantity}

We conduct ablation studies to validate the effectiveness of each proposed component (more ablations and stress tests are provided in the Appendix).  
As shown in Table~\ref{tab:abla_quant_metrics}, removing any individual module leads to a noticeable drop in all metrics.  
In addition to quantitative metrics, we also visualize representative ablation results with details as follows.

\paragraph{Semantic Feature Enhancement}

Semantic Feature Enhancement improves the low spatial resolution of DINOv3 features. 
Without it (Fig.~\ref{fig:ablation_1}), noisy projected features lead to unstable clustering and fragmented component retrieval. 
Though the target region can still be roughly localized, the noisy 3D masks introduce geometric artifacts during inpainting, such as unintended structure growth in empty areas.

\paragraph{3D Component Cluster \& Retrieval}

The 3D Component Cluster \& Retrieval module enables accurate localization from semantic queries. 
As shown in Fig.~\ref{fig:ablation_1}, directly selecting voxels using top-$k$ cosine similarity requires manual tuning and is sensitive to feature noise. 
This often retrieves incomplete regions; for example, only part of a wheel is selected, causing the edited result to mix the original and newly generated appearances.

\vspace{-0.5em}
\paragraph{Refined Inpainting Mask}
The refined inpainting mask not only prevents the model from modifying regions that should remain unchanged, but also ensures that the edited component can grow naturally within a controlled spatial boundary. 
As shown in Fig.~\ref{fig:ablation_2}, removing this refinement causes the inpainting process to overwrite preserved areas and regenerate unintended geometry: the new shield bleeds into the torso, and the sword, which is supposed to remain unchanged, gets regenerated, resulting in the character holding two swords. 
These artifacts demonstrate the necessity of mask refinement for stable and localized 3D editing.

\vspace{-0.5em}
\paragraph{Stacked Image}
The stacked image provides spatial context for local reference images. 
Without stacking, the encoder receives only a local patch without positional cues, making it difficult to align the reference with the global 3D asset. 
As shown in Fig.~\ref{fig:ablation_2}, this leads to misaligned and geometrically inconsistent edits, whereas stacking provides spatial alignment and improves both semantic and geometric consistency.

%% file: tables/compare_quantity.tex
\begin{table}[t]
\centering
\scriptsize
\vspace{-0.5em}
\setlength{\tabcolsep}{1.5pt}
\resizebox{\columnwidth}{!}{%
\begin{tabular}{@{}lcccc@{}}
\toprule
\textbf{Metric} & \textbf{Ours} & \textbf{IP-Adapter} & 
\textbf{TRELLIS} & \textbf{Fuse3D}\\
\midrule
CLIP $\uparrow$ &\textbf{32.19}  & 29.02 &29.38  & 30.29 \\
Reg.-Spec. CLIP $\uparrow$ & \textbf{77.94} & 73.63 &72.61  & 76.16  \\
ImageReward $\uparrow$ & \textbf{0.5853} & 0.0198 &0.2501  &0.1738  \\
CD$_{\text{keep}}$ ($\times 10^{-3}$) $\downarrow$ & \textbf{0.91} & 13.12 & 11.25 & 9.87 \\
3D BBox IoU $\uparrow$ & \textbf{0.73} & N/A & N/A & 0.39 \\
\bottomrule
\end{tabular}
}
\caption{Quantitative comparison of different methods.}
\label{tab:quant_metrics}
\end{table}

%% file: tables/gpt_eval.tex
\begin{table}
\centering
\setlength{\tabcolsep}{2.6mm}
\small
\begin{tabular*}{\columnwidth}{@{\extracolsep{\fill}}lccc@{}}
\toprule
\textbf{Method} & \textbf{V. Q.} & \textbf{E. A.} & \textbf{P. C.}
\\ 
\hline
vs. Fuse3D & 59 & 53 & 67   \\
vs. IP-Adapter & 98 & 92 & 91  \\
vs. TRELLIS & 62 & 57 &  53 \\
\bottomrule
\end{tabular*}
\caption{
GPTEvals3D results (\% preferred) for 3D editing across different methods. Scores above 50 indicate that our method is preferred.
}
\label{tab:GPTcompare}
\end{table}

%% file: tables/userStudy.tex
\begin{table}
\centering
\setlength{\tabcolsep}{1.3mm}
\begin{tabular*}{\columnwidth}{@{\extracolsep{\fill}}lccc@{}}
\toprule
\textbf{Method} & \textbf{V. Q.} & \textbf{E. A.} & \textbf{P. C.}
\\ 
\hline
Ours & \textbf{4.4} & \textbf{4.6} & \textbf{4.8}   \\
Fuse3D & 4.1 & 3.9 & 4.1   \\
IP-Adapter & 2.8 & 3.1 & 2.1  \\
TRELLIS & 3.9 & 3.3 & 4.7   \\
\bottomrule
\end{tabular*}
\caption{
User ratings (1--5) for different methods. 
}
\label{tab:userStudy}
\end{table}

%% file: tables/ablation_quantity.tex
\begin{table}[t]
\centering
\scriptsize
\setlength{\tabcolsep}{4pt}
\renewcommand{\arraystretch}{1.05}
\vspace{-0.5em}
\resizebox{\columnwidth}{!}{%
\begin{tabular}{lccc}
\toprule
\textbf{Method} & \textbf{CLIP $\uparrow$} & \textbf{Reg.-Spec. CLIP $\uparrow$} & \textbf{ImageReward $\uparrow$} \\
\midrule
\textbf{Ours} & \textbf{32.19} & \textbf{77.94} & \textbf{0.5853} \\
w/o Semantic Feature Enhancement & 30.02 & 72.10 & 0.2399 \\
w/o 3D Component Cluster Retrieval & 30.15 & 71.40 & 0.1764 \\
w/o Refined Inpainting Mask & 30.78 & 76.23 & 0.3264 \\
w/o 3D Stacked Image & 27.33 & 72.53 & 0.2291 \\
\bottomrule
\end{tabular}
}
\caption{Quantitative comparison of ablation study.}
\label{tab:abla_quant_metrics}
\end{table}

%% file: sec/5_misc.tex
\section{Conclusion}


We presented \name, a framework for component-aware 3D editing that integrates semantic understanding directly into 3D space. 
To enable reliable semantic retrieval, we construct a 3D semantic embedding by encoding multi-view renderings with a semantically aligned image encoder, enhanced through a semantic feature enhancement strategy, and projecting the resulting features onto the voxelized surface of the asset. 
Clustering this embedding allows us to robustly retrieve editable 3D components using either local reference images or optional text queries.
For editing, we first construct unified condition tokens, followed by a two-stage inpainting procedure within a pretrained 3D generative model. 
This design enables ES3D to perform precise edits while preserving unedited regions.

Extensive experiments show that ES3D delivers more accurate retrieval, stronger geometric editing, and higher visual quality than existing 3D editing methods. We believe this work moves toward more controllable and user-friendly 3D asset editing.

%% file: aaai2027.bib
@inproceedings{meshEditing,
      title={3D Mesh Editing using Masked LRMs}, 
      author={Gao, William and Wang, Dilin and Fan, Yuchen and Božič, Aljaž and Stuyck, Tuur and Li, Zhengqin and Dong, Zhao and Ranjan, Rakesh and Sarafianos, Nikolaos},
      booktitle={ICCV},
      year={2025}
    }

@inproceedings{
progressive3D,
title={Progressive3D: Progressively Local Editing for Text-to-3D Content Creation with Complex Semantic Prompts},
author={Xinhua Cheng and Tianyu Yang and Jianan Wang and Yu Li and Lei Zhang and Jian Zhang and Li Yuan},
booktitle={The Twelfth International Conference on Learning Representations},
year={2024},
url={https://openreview.net/forum?id=O072Rc8uUy}
}

@inproceedings{
A3D,
title={A3D: Does Diffusion Dream about 3D Alignment?},
author={Savva Victorovich Ignatyev and Nina Konovalova and Daniil Selikhanovych and Oleg Voynov and Nikolay Patakin and Ilya Olkov and Dmitry Senushkin and Alexey Artemov and Anton Konushin and Alexander Filippov and Peter Wonka and Evgeny Burnaev},
booktitle={The Thirteenth International Conference on Learning Representations},
year={2025},
url={https://openreview.net/forum?id=QQCIfkhGIq}
}

@misc{focalDream,
      title={FocalDreamer: Text-driven 3D Editing via Focal-fusion Assembly}, 
      author={Yuhan Li and Yishun Dou and Yue Shi and Yu Lei and Xuanhong Chen and Yi Zhang and Peng Zhou and Bingbing Ni},
      year={2023},
      eprint={2308.10608},
      archivePrefix={arXiv},
      primaryClass={cs.CV},
      url={https://arxiv.org/abs/2308.10608}, 
}

@misc{dinov3,
  title={{DINOv3}},
  author={Sim{\'e}oni, Oriane and Vo, Huy V. and Seitzer, Maximilian and Baldassarre, Federico and Oquab, Maxime and Jose, Cijo and Khalidov, Vasil and Szafraniec, Marc and Yi, Seungeun and Ramamonjisoa, Micha{\"e}l and Massa, Francisco and Haziza, Daniel and Wehrstedt, Luca and Wang, Jianyuan and Darcet, Timoth{\'e}e and Moutakanni, Th{\'e}o and Sentana, Leonel and Roberts, Claire and Vedaldi, Andrea and Tolan, Jamie and Brandt, John and Couprie, Camille and Mairal, Julien and J{\'e}gou, Herv{\'e} and Labatut, Patrick and Bojanowski, Piotr},
  year={2025},
  eprint={2508.10104},
  archivePrefix={arXiv},
  primaryClass={cs.CV},
  url={https://arxiv.org/abs/2508.10104},
}

@inproceedings{multi_crop_train,
author = {Caron, Mathilde and Misra, Ishan and Mairal, Julien and Goyal, Priya and Bojanowski, Piotr and Joulin, Armand},
title = {Unsupervised learning of visual features by contrasting cluster assignments},
year = {2020},
isbn = {9781713829546},
publisher = {Curran Associates Inc.},
address = {Red Hook, NY, USA},
booktitle = {Proceedings of the 34th International Conference on Neural Information Processing Systems},
articleno = {831},
numpages = {13},
location = {Vancouver, BC, Canada},
series = {NIPS '20}
}

@misc{dino,
      title={Emerging Properties in Self-Supervised Vision Transformers}, 
      author={Mathilde Caron and Hugo Touvron and Ishan Misra and Hervé Jégou and Julien Mairal and Piotr Bojanowski and Armand Joulin},
      year={2021},
      eprint={2104.14294},
      archivePrefix={arXiv},
      primaryClass={cs.CV},
      url={https://arxiv.org/abs/2104.14294}, 
}

@misc{repaint,
      title={RePaint: Inpainting using Denoising Diffusion Probabilistic Models}, 
      author={Andreas Lugmayr and Martin Danelljan and Andres Romero and Fisher Yu and Radu Timofte and Luc Van Gool},
      year={2022},
      eprint={2201.09865},
      archivePrefix={arXiv},
      primaryClass={cs.CV},
      url={https://arxiv.org/abs/2201.09865}, 
}

@InProceedings{nullText,
    author    = {Mokady, Ron and Hertz, Amir and Aberman, Kfir and Pritch, Yael and Cohen-Or, Daniel},
    title     = {NULL-Text Inversion for Editing Real Images Using Guided Diffusion Models},
    booktitle = {Proceedings of the IEEE/CVF Conference on Computer Vision and Pattern Recognition (CVPR)},
    month     = {June},
    year      = {2023},
    pages     = {6038-6047}
}

@article{p2p,
  title={Prompt-to-prompt image editing with cross attention control},
  author={Hertz, Amir and Mokady, Ron and Tenenbaum, Jay and Aberman, Kfir and Pritch, Yael and Cohen-Or, Daniel},
  booktitle={arXiv preprint arXiv:2208.01626},
  year={2022}
}

@InProceedings{plugPlay,
    author    = {Tumanyan, Narek and Geyer, Michal and Bagon, Shai and Dekel, Tali},
    title     = {Plug-and-Play Diffusion Features for Text-Driven Image-to-Image Translation},
    booktitle = {Proceedings of the IEEE/CVF Conference on Computer Vision and Pattern Recognition (CVPR)},
    month     = {June},
    year      = {2023},
    pages     = {1921-1930}
}

@inproceedings{
      sdedit,
      title={{SDE}dit: Guided Image Synthesis and Editing with Stochastic Differential Equations},
      author={Chenlin Meng and Yutong He and Yang Song and Jiaming Song and Jiajun Wu and Jun-Yan Zhu and Stefano Ermon},
      booktitle={International Conference on Learning Representations},
      year={2022},
}

@misc{pointE,
      title={Point-E: A System for Generating 3D Point Clouds from Complex Prompts}, 
      author={Alex Nichol and Heewoo Jun and Prafulla Dhariwal and Pamela Mishkin and Mark Chen},
      year={2022},
      eprint={2212.08751},
      archivePrefix={arXiv},
      primaryClass={cs.CV},
      url={https://arxiv.org/abs/2212.08751}, 
}

@inproceedings{SDFusion,
  author={Cheng, Yen-Chi and Lee, Hsin-Ying and Tuyakov, Sergey and Schwing, Alex and Gui, Liangyan},
  title={{SDFusion}: Multimodal 3D Shape Completion, Reconstruction, and Generation},
  booktitle={CVPR},
  year={2023},
}

@misc{shapeE,
      title={Shap-E: Generating Conditional 3D Implicit Functions}, 
      author={Heewoo Jun and Alex Nichol},
      year={2023},
      eprint={2305.02463},
      archivePrefix={arXiv},
      primaryClass={cs.CV},
      url={https://arxiv.org/abs/2305.02463}, 
}

@inproceedings{diffusionSDF,
  author={Li, Muheng and Duan, Yueqi and Zhou, Jie and Lu, Jiwen},
  title={Diffusion-SDF: Text-to-Shape via Voxelized Diffusion},
  booktitle={Proceedings of the IEEE Conference on Computer Vision and Pattern Recognition (CVPR)},
  year={2023}
}

@inproceedings{GET3D,
title={GET3D: A Generative Model of High Quality 3D Textured Shapes Learned from Images},
author={Jun Gao and Tianchang Shen and Zian Wang and Wenzheng Chen and Kangxue Yin
and Daiqing Li and Or Litany and Zan Gojcic and Sanja Fidler},
booktitle={Advances In Neural Information Processing Systems},
year={2022}
}

@inproceedings{
lrm,
title={{LRM}: Large Reconstruction Model for Single Image to 3D},
author={Yicong Hong and Kai Zhang and Jiuxiang Gu and Sai Bi and Yang Zhou and Difan Liu and Feng Liu and Kalyan Sunkavalli and Trung Bui and Hao Tan},
booktitle={The Twelfth International Conference on Learning Representations},
year={2024},
url={https://openreview.net/forum?id=sllU8vvsFF}
}

@misc{dreamFusion,
      title={DreamFusion: Text-to-3D using 2D Diffusion}, 
      author={Ben Poole and Ajay Jain and Jonathan T. Barron and Ben Mildenhall},
      year={2022},
      eprint={2209.14988},
      archivePrefix={arXiv},
      primaryClass={cs.CV},
      url={https://arxiv.org/abs/2209.14988}, 
}

@misc{lgm,
      title={LGM: Large Multi-View Gaussian Model for High-Resolution 3D Content Creation}, 
      author={Jiaxiang Tang and Zhaoxi Chen and Xiaokang Chen and Tengfei Wang and Gang Zeng and Ziwei Liu},
      year={2024},
      eprint={2402.05054},
      archivePrefix={arXiv},
      primaryClass={cs.CV},
      url={https://arxiv.org/abs/2402.05054}, 
}

@article{ldm,
  title={LDM: Large Tensorial SDF Model for Textured Mesh Generation},
  author={Xie, Rengan and Zheng, Wenting and Huang, Kai and Chen, Yizheng and Wang, Qi and Ye, Qi and Chen, Wei and Huo, Yuchi},
  journal={arXiv preprint arXiv:2405.14580},
  year={2024}
}

@InProceedings{trellis,
    author    = {Xiang, Jianfeng and Lv, Zelong and Xu, Sicheng and Deng, Yu and Wang, Ruicheng and Zhang, Bowen and Chen, Dong and Tong, Xin and Yang, Jiaolong},
    title     = {Structured 3D Latents for Scalable and Versatile 3D Generation},
    booktitle = {Proceedings of the Computer Vision and Pattern Recognition Conference (CVPR)},
    month     = {June},
    year      = {2025},
    pages     = {21469-21480}
}

@misc{ultra3d,
  title={Ultra3D: Efficient and High-Fidelity 3D Generation with Part Attention},
  author={Yiwen Chen and Zhihao Li and Yikai Wang and Hu Zhang and Qin Li and Chi Zhang and Guosheng Lin},
  year={2025},
  eprint={2507.17745},
  archivePrefix={arXiv},
  primaryClass={cs.CV},
  url={https://arxiv.org/abs/2507.17745}
}

@article{direct3d2s,
  title={Direct3D-S2: Gigascale 3D Generation Made Easy with Spatial Sparse Attention}, 
  author={Shuang Wu and Youtian Lin and Feihu Zhang and Yifei Zeng and Yikang Yang and Yajie Bao and Jiachen Qian and Siyu Zhu and Philip Torr and Xun Cao and Yao Yao},
  journal={arXiv preprint arXiv:2505.17412},
  year={2025}
}

@inproceedings{instructNerf,
    author = {Haque, Ayaan and Tancik, Matthew and Efros, Alexei and Holynski, Aleksander and Kanazawa, Angjoo},
    title = {Instruct-NeRF2NeRF: Editing 3D Scenes with Instructions},
    booktitle = {Proceedings of the IEEE/CVF International Conference on Computer Vision},
    year = {2023},
}

@inproceedings{flow,
title={Flow Matching for Generative Modeling},
author={Yaron Lipman and Ricky T. Q. Chen and Heli Ben-Hamu and Maximilian Nickel and Matthew Le},
booktitle={The Eleventh International Conference on Learning Representations },
year={2023},
url={https://openreview.net/forum?id=PqvMRDCJT9t}
}

@misc{ipAdapter,
      title={IP-Adapter: Text Compatible Image Prompt Adapter for Text-to-Image Diffusion Models}, 
      author={Hu Ye and Jun Zhang and Sibo Liu and Xiao Han and Wei Yang},
      year={2023},
      eprint={2308.06721},
      archivePrefix={arXiv},
      primaryClass={cs.CV},
      url={https://arxiv.org/abs/2308.06721}, 
}

@article{pro3d,
        title={Pro3D-Editor: A Progressive-Views Perspective for Consistent and Precise 3D Editing},
        author={Zheng, Yang and Huang, Mengqi and Chen, Nan and Mao, Zhendong},
        journal={arXiv preprint arXiv:2506.00512},
        year={2025}
}

@inproceedings{gptEval,
   author = {Tong Wu and Guandao Yang and Zhibing Li and Kai Zhang and
             Ziwei Liu and Leonidas Guibas and Dahua Lin and Gordon Wetzstein},
   title = {GPT-4V(ision) is a Human-Aligned Evaluator for Text-to-3D Generation},
   booktitle = {CVPR},
   year = {2024},
}

@InProceedings{blendedDiffusion,
        author    = {Avrahami, Omri and Lischinski, Dani and Fried, Ohad},
        title     = {Blended Diffusion for Text-Driven Editing of Natural Images},
        booktitle = {Proceedings of the IEEE/CVF Conference on Computer Vision and Pattern Recognition (CVPR)},
        month     = {June},
        year      = {2022},
        pages     = {18208-18218}
}

@article{glide,
  title={Glide: Towards photorealistic image generation and editing with text-guided diffusion models},
  author={Nichol, Alex and Dhariwal, Prafulla and Ramesh, Aditya and Shyam, Pranav and Mishkin, Pamela and McGrew, Bob and Sutskever, Ilya and Chen, Mark},
  journal={arXiv preprint arXiv:2112.10741},
  year={2021}
}

@inproceedings{ledit++,
  title={Ledits++: Limitless image editing using text-to-image models},
  author={Brack, Manuel and Friedrich, Felix and Kornmeier, Katharia and Tsaban, Linoy and Schramowski, Patrick and Kersting, Kristian and Passos, Apolin{\'a}rio},
  booktitle={Proceedings of the IEEE/CVF conference on computer vision and pattern recognition},
  pages={8861--8870},
  year={2024}
}

@inproceedings{turboEdit,
  title={Turboedit: Text-based image editing using few-step diffusion models},
  author={Deutch, Gilad and Gal, Rinon and Garibi, Daniel and Patashnik, Or and Cohen-Or, Daniel},
  booktitle={SIGGRAPH Asia 2024 Conference Papers},
  pages={1--12},
  year={2024}
}

@InProceedings{masactrl,
    author    = {Cao, Mingdeng and Wang, Xintao and Qi, Zhongang and Shan, Ying and Qie, Xiaohu and Zheng, Yinqiang},
    title     = {MasaCtrl: Tuning-Free Mutual Self-Attention Control for Consistent Image Synthesis and Editing},
    booktitle = {Proceedings of the IEEE/CVF International Conference on Computer Vision (ICCV)},
    month     = {October},
    year      = {2023},
    pages     = {22560-22570}
}

@inproceedings{fuse3d,
author = {Jin, Xuancheng and Xie, Rengan and Zheng, Wenting and Wang, Rui and Bao, Hujun and Huo, Yuchi},
title = {Fuse3D: Generating 3D Assets Controlled by Multi-Image Fusion},
year = {2025},
isbn = {9798400721373},
publisher = {Association for Computing Machinery},
address = {New York, NY, USA},
url = {https://doi.org/10.1145/3757377.3763943},
doi = {10.1145/3757377.3763943},
booktitle = {Proceedings of the SIGGRAPH Asia 2025 Conference Papers},
articleno = {144},
numpages = {12},
location = {
},
series = {SA Conference Papers '25}
}

@misc{nano3d,
      title={NANO3D: A Training-Free Approach for Efficient 3D Editing Without Masks}, 
      author={Junliang Ye and Shenghao Xie and Ruowen Zhao and Zhengyi Wang and Hongyu Yan and Wenqiang Zu and Lei Ma and Jun Zhu},
      year={2025},
      eprint={2510.15019},
      archivePrefix={arXiv},
      primaryClass={cs.CV},
      url={https://arxiv.org/abs/2510.15019}, 
}

@inproceedings{preditor3d,
  author={Ziya Erkoç and Can Gümeli and Chaoyang Wang and Matthias Nießner and Angela Dai and Peter Wonka and Hsin-Ying Lee and Peiye Zhuang},
  title={PrEditor3D: Fast and Precise 3D Shape Editing},
  year={2025},
  cdate={1735689600000},
  pages={640-649},
  url={https://openaccess.thecvf.com/content/CVPR2025/html/Erkoc_PrEditor3D_Fast_and_Precise_3D_Shape_Editing_CVPR_2025_paper.html},
  booktitle={CVPR},
}

@misc{vae,
      title={Auto-Encoding Variational Bayes}, 
      author={Diederik P Kingma and Max Welling},
      year={2022},
      eprint={1312.6114},
      archivePrefix={arXiv},
      primaryClass={stat.ML},
      url={https://arxiv.org/abs/1312.6114}, 
}

@article{ddpm,
    title={Denoising Diffusion Probabilistic Models},
    author={Jonathan Ho and Ajay Jain and Pieter Abbeel},
    year={2020},
    journal={arXiv preprint arxiv:2006.11239}
}

@article{MagicClay,
title = {MagicClay: Sculpting Meshes With Generative Neural Fields},
author = {Amir Barda and Vladimir G. Kim and Noam Aigerman and Amit H. Bermano and Thibault Groueix},
year = {2024},
journal = {SIGGRAPH Asia (Conference track)}}

@misc{Instant3dit,
      title={Instant3dit: Multiview Inpainting for Fast Editing of 3D Objects}, 
      author={Amir Barda and Matheus Gadelha and Vladimir G. Kim and Noam Aigerman and Amit H. Bermano and Thibault Groueix},
      journal = {IEEE Conference on Computer Vision and Pattern Recognition (CVPR)},
      year = {2025}, 
}

@misc{pixbrush,
      title={3D PixBrush: Image-Guided Local Texture Synthesis}, 
      author={Dale Decatur and Itai Lang and Kfir Aberman and Rana Hanocka},
      year={2025},
      eprint={2507.03731},
      archivePrefix={arXiv},
      primaryClass={cs.GR},
      url={https://arxiv.org/abs/2507.03731}, 
}
